\documentclass[12pt]{article}

\usepackage[T1]{fontenc}
\usepackage[utf8]{inputenc}
\usepackage[english]{babel}

\usepackage{siunitx}

\usepackage{physics}

\usepackage{tkz-graph}
\usepackage{tikz}
\usetikzlibrary{arrows,matrix,decorations.pathreplacing,positioning,chains,fit,shapes,calc} 
\usetikzlibrary{calc,patterns,decorations.pathmorphing,decorations.markings}

\usepackage{libertine}
\usepackage{libertinust1math}

\usepackage{tikz-qtree}

\usepackage{amsmath}
\usepackage{amssymb}
\usepackage{bm}
\usepackage{bbm}
\usepackage{spalign}

\newcommand{\BlackBox}{\rule{1.5ex}{1.5ex}}  
\newenvironment{proof}{\par\noindent{\bf Proof\ }}{\hfill\BlackBox\\[2mm]}

\newtheorem{proposition}{Proposition} 

\usepackage[obeyspaces]{url}

\usepackage{graphicx}
\graphicspath{{images/}}
\usepackage{float}

\usepackage{fancyvrb}
\usepackage{empheq}

\usepackage{jlcode}
\addlitjlmacros{@constraint}{@constraint}{11}
\addlitjlmacros{@variable}{@variable}{9}
\addlitjlmacros{@objective}{@objective}{10}
\addlitjlmacros{@functor}{@functor}{8}

\usepackage{todonotes}
\usepackage{subcaption}
\usepackage{booktabs}

\usepackage{breakcites}
\usepackage[htt]{hyphenat}
\usepackage{jmlr2e}
\usepackage[scaled=0.85]{FiraMono}

\usepackage[capitalize,nameinlink,noabbrev]{cleveref}[0.21]

\DeclareMathOperator*{\argmin}{argmin}
\DeclareMathOperator*{\sign}{sign}
\DeclareMathOperator*{\triu}{triu}

\newcommand{\Reals}{{\ensuremath{\mathbb{R}}}}

\definecolor{brandeisblue}{rgb}{0.0, 0.44, 1.0}

\usepackage{natbib}
\setcitestyle{authoryear,round,citesep={;},aysep={,},yysep={;}}

\title{Flexible Differentiable Optimization via Model Transformations}

\date{}

\author{\noindent
\name Mathieu Besançon\thanks{The authors are ordered alphabetically and not by order of contribution}
\email \href{mailto:besancon@zib.de}{besancon@zib.de} \\
       \addr Zuse Institute Berlin, Germany
       \AND
\name Joaquim Dias Garcia \email  \href{mailto:joaquim@psr-inc.com}{joaquim@psr-inc.com}\\
\addr PSR \& PUC-Rio, Rio de Janeiro, Brazil
\AND
\name Benoît Legat \email \href{mailto:blegat@mit.edu}{blegat@mit.edu} \\
\addr Massachusetts Institute of Technology, MA, USA
\AND
\name Akshay Sharma \email \href{mailto:akshay.s@columbia.edu }{akshay.s@columbia.edu} \\
\addr Columbia University, NY, USA
}

\begin{document}

\maketitle

\abstract{
We introduce DiffOpt.jl, a Julia library to differentiate through the solution
of optimization problems with respect to arbitrary parameters present in
the objective and/or constraints. The library builds upon MathOptInterface,
thus leveraging the rich ecosystem of solvers and composing 
well with modeling languages like JuMP.
DiffOpt offers both forward and reverse differentiation modes, enabling multiple
use cases from hyperparameter optimization to backpropagation and sensitivity
analysis, bridging constrained optimization with end-to-end differentiable 
programming.
DiffOpt is built on two known rules for differentiating quadratic programming and conic programming standard forms.
However, thanks ability to differentiate through model transformation,
the user is not limited to these forms and can differentiate with respect to the parameters of any model that can be reformulated into these standard forms.
This notably includes programs mixing affine conic constraints and convex quadratic constraints or objective function.

}

\vspace*{10pt}

\section{Introduction}

Differentiable Optimization ($\partial \mathcal{O}$) has become a center
of interest in the last years, both as a stand-alone methodology
providing additional information on the sensitivity of generic optimization problems
and as a principled way to integrate complex convex optimization components in Machine Learning,
following a broader trend of \emph{differentiable programming} which extends the set of computer programs
for which derivatives can be computed \citep{innes2019differentiable}.
In that context, $\partial \mathcal{O}$ can be seen as a form of differentiable programming, where the procedure for which derivatives are computed is the algorithm solving an optimization problem.
We present DiffOpt, a library implementing key methods for $\partial \mathcal{O}$
in Julia \citep{bezanson2017julia} based on the \texttt{MathOptInterface} (MOI) abstract data structure~\citep{legat2020mathoptinterface}. Consequently, the extensions to the usual mathematical optimization machinery are also available in \texttt{JuMP} \citep{DunningHuchetteLubin2017} with MOI as its backend. The library associated with this paper is archived at \citet{DiffOptArchive}.

$\partial \mathcal{O}$ regroups a set of methods to compute derivative information
of a function that includes an optimization problem. We start considering a generic finite-dimensional optimization problem
parameterized by $\mathbf{\theta}$:
\begin{align*}
\mathcal{P}(\theta): \min_{x}\,\,\, &  G_0(x,\theta) & \\
\text{s.t.}\,\,\, & G_i(x, \theta) \in \mathcal{S}_i \,\,\, \forall i \in \{1,\dots,m\}, &
\end{align*}
where $x \subseteq \Reals^n$, $\theta \in \Theta \subseteq \Reals^k $, $G_i$ are functions $\Reals^n\times \Theta\rightarrow \Reals^{n_i}$ and $\mathcal{S}_i \subset \Reals^{n_i}$.
For each $\theta$ we define the solution map $\Psi(\theta) = \argmin_{x} \mathcal{P}(\theta)$, a set-valued function.
We will focus on the case where the solution map is single-valued: $\psi(\theta) = x^*(\theta) = \argmin_{x} \mathcal{P}(\theta)$.

Differentiable optimization in its generic form entails computing the
derivative of the output $x^*(\theta)$ with
respect to all parameters, i.e., computing the Jacobian matrix:
\begin{equation*}
    \frac{\partial x^*}{\partial \theta} \in \Reals^{k\times n}.
\end{equation*}
\noindent
Instead of the Jacobian matrix of the solution map, we often reason on the derivative of the solution map as a linear map denoted as
$D \psi(\theta)$, with adjoint $D^\top \psi(\theta)$. In the general case, in which set-valued functions are considered, more sophisticated objects are required, such as the Generalized Jacobian \citep{dempe2001generalized, stechlinski2018generalized}, which are based on the seminal works on sensitivity analysis by \citet{fiacco1983introduction} and \citet{robinson1982generalized}.
On the other hand, even in the single-valued case, the solution is not smooth in general, which poses further challenges to the derivative computation process.
Some lines of work have investigated non-smooth automatic (sub)differentiation and the behavior
of automatic differentiation in the non-smooth case
\citep{kakade2018provably,bolte2020mathematical}.
However, in most cases in differentiable programming, a heuristic quantity is computed when the limit
of the directional derivative is ill-defined\footnote{We point the interested reader to the ChainRules.jl recommendations \url{https://juliadiff.org/ChainRulesCore.jl/v1.15/maths/nondiff_points.html}, accessed March 2023}.

The recent survey \citet{kotary2021end} on learning with constrained
optimization offers a review of $\partial \mathcal{O}$ techniques in light of the recent trends of combining optimization and machine learning.
However, the recent works on $\partial \mathcal{O}$ have been using many techniques developed for sensitivity analysis of optimization problems. For more information on previous works regarding the latter, the reader is directed to \citet{fiacco1983introduction,fiacco1990sensitivity,gal2010postoptimal,bonnans2013perturbation}.

For differentiation with respect to arbitrary problem parameters, two main methods
have recently appeared in the literature to handle structured problems.
The first method, from \citet{amos_optnet_2019}, can be
applied to convex quadratic optimization problems with linear equality and inequality constraints.
While the second, from 
\citet{agrawal_differentiating_2020}, is applied to conic optimization problems.
These two methods are implemented in \texttt{DiffOpt.jl} as \emph{differentiation rules} described in \cref{sec:rules}.
These rules are then combined with model transformations in order to communicate the sensitivity information to the form used by the user.
As detailed in \cref{sec:quad_to_soc},
this notably enables the users to obtain the sensitivity with respect to quadratic constraints in a model mixing quadratic constraints and affine conic constraints.

We briefly go over the lines of work that
are closest to the methods of \citet{amos_optnet_2019,agrawal_differentiating_2020}.\\

A part of the recent work on $\partial \mathcal{O}$ has focused on specific 
optimization problems.
In \citet{blondel_fast_2020}, the authors consider a differentiable optimization 
method for sorting and ranking problems.
In \citet{berthet_learning_2020}, the differentiation of the solution to a generic 
convex optimization problem with respect to a linear objective is considered.
When the feasible set is a polytope,
the issue of a Jacobian matrix being zero almost everywhere arises similarly
to the ranking problem from \citet{blondel_fast_2020}.
The approach followed is that of perturbed optimization,
considering the input cost vector as a random variable centered around a
nominal value. This allows sampling the output solution and Jacobian matrix
from input cost vectors generated from the distribution,
yielding unbiased estimators for both while only requiring access to a linear 
minimization oracle, i.e., \ duality information is not required.
That line of work on differentiation through stochastic perturbation has been extended
in \citet{dalle2022learning}, which offers a unified toolbox for stochastic
differentiation with respect to objective parameters.

The methodology of differentiable conic problems has been extended to log-log convex
problems in \citet{agrawal_loglog_2020}, using the grammar from Disciplined Geometric Programming
and the rules established for parameterized disciplined convex problems in \citet{agrawal_differentiable_2019}.
A differentiable method has been developed for submodular functions in \citet{djolonga2018differentiable},
opening $\partial \mathcal{O}$ to optimization problems with discrete structures.
In \citet{gould2019deep}, deep learning models are studied with nonlinear optimization problems
as nodes instead of closed-form functions, defining a differentiable optimization method
for nonlinear problems without requiring convexity.
The authors leverage implicit differentiation of the Lagrangian reformulation at the optimal point
to estimate derivative information of the output solution with respect to the input parameters of the node.
The framework proposed in \citet{paulus2021comboptnet} extends
$\partial O$ to optimization problems including integer constraints. 
Similar to previous work in the convex setting and unlike prior models tackling combinatorial problems, their method handles the differentiation of constraints
using an estimation of active constraints at the optimum in the backpropagation phase.

In \citet{blondel2021efficient}, differentiable optimization
is viewed as a problem of implicit differentiation.
By expressing the solution map of the optimization problem as the root of a system of equations (such as the KKT conditions)
or the solution to a fixed point equation, the system can
leverage automatic differentiation (AD) of the implicit equations to differentiate the solution map.
The approach generalizes the differentiation of quadratic and conic optimization from \citet{amos_optnet_2019} and \citet{agrawal_differentiable_2019} but requires a user-defined system of equations that are necessary for optimality.\\

The rest of this paper is structured as follows.
\cref{sec:background} covers the background and related research on
$\partial \mathcal{O}$.
\cref{sec:structure} presents the structure of the DiffOpt package and important
features. \cref{sec:examples} highlights some applications illustrating the use
of the package.

\section{Differentiating Convex Optimization Problems}\label{sec:background}

We focus now on the theoretical background of the two methods implemented in \texttt{DiffOpt.jl}.
First, we detail the convex quadratic optimization case \citep{amos_optnet_2019}, and, in the sequence, we explain the conic optimization case 
\citep{agrawal_differentiating_2020}.
Both methods are based on rewriting necessary optimality conditions for the optimization problem as a system of nonlinear equations that will locally define the primal-dual solution as an implicit function and then apply the Implicit Function Theorem, IFT, \citep[Theorem 1B.1, page 17]{dontchev2009implicit}, to obtain the derivative of such implicit function. This is in line with early methods from \citet{fiacco1990nonlinear}. We highlight that other versions of the IFT can be used if required.

The above mentioned IFT from \cite{dontchev2009implicit} is as follows:

\noindent
\textbf{Implicit Function Theorem}{ \citep{dontchev2009implicit}.
Let $f : \mathbb{R}^d \times \mathbb{R}^n \rightarrow \mathbb{R}^n$ be continuously differentiable in a neighborhood of $(p^*,x^*)$ and such that $f(p^*,x^*) = 0$. Let the partial Jacobian of $f$ with respect to $x$ at $(p^*,x^*)$, namely $D_x f(p^*,x^*)$, be nonsingular. Then the solution mapping $S(p) = \{x\in \Reals^n \,|\, f(p, x)=0\}$ for $p\in \Reals^d$ has a single-valued localization $s$ around $p^*$ for $x^*$, hence, implicitly defining the function
\begin{align*}
s : \mathbb{R}^d \rightarrow \mathbb{R}^n, \quad s(p) = x,
\end{align*}
which is continuously differentiable in a neighborhood $Q$ of $p^*$ with a Jacobian satisfying:
\begin{align*}
D s(p) = -D_x f(p,s(p))^{-1} D_p f(p,s(p))\,\,\, \forall p \in Q.
\end{align*}
}

\subsection{Quadratic programs}
\label{sec:qp}

A major step towards integrating differentiable optimization into ML pipelines
was presented in \citet{amos_optnet_2019,amos_differentiable_2019} for convex 
quadratic problems (QP). The considered optimization models are of the form:
\begin{align*}
 \min_{x \in \Reals^n}\,\, & \frac12 x^\top Q x + c^\top x \\
& G x \leq h : (\lambda) \\
& A x = b : (\mu)
\end{align*}
where $\lambda \in \Reals^p$, $\mu \in \Reals^m$ denote the dual variables associated
with the inequality and equality constraints, respectively.
Unlike prior work, the solution map is differentiated with respect to all
problem data 
$(A \in \Reals^{m \times n}, b \in \Reals^m, G \in \Reals^{p \times n}, h \in \Reals^p, Q\in \mathbb{S}^{n}_{+}, c \in \Reals^n)$.
In particular, differentiating the solution with respect to constraint 
coefficients open new applications, including learning the constraints of
the convex problem along with solutions as illustrated in \citet{amos_differentiable_2019}
on generic polytopes and combinatorial problems.
The differentiation method starts by representing the solution process as solving
a system of equations.
The solution can then be differentiated with respect to its parameters using 
implicit differentiation.
In the case of QPs, the KKT conditions of the system fully describe the
optimality conditions for a primal-dual solution:
\begin{align*}
& Qx + c + A^\top \mu + G^\top \lambda = 0 && (\nabla L) \\
& Ax = b && (P_{eq})\\
& 0 \leq h - G x\,\, \bot \,\, \lambda \geq 0 && (C),
\end{align*}
where $(\nabla L)$ represents the gradient of the Lagrangian, $(P_{eq})$ represents the
primal feasibility of equality constraints and $(C)$ includes primal feasibility of inequality constraints, 
complementarity and dual feasibility for inequalities.
A system of equality constraints can be derived to compute derivatives:
\begin{align*}
& (\nabla  L),\,\, (P_{eq}), \\
& \lambda_i (h - Gx)_i = 0 \,\,\,\forall i.
\end{align*}
This representation is not equivalent to the KKT conditions but is a sufficient set of equations to compute sensitivities at a given primal (and dual) solution, as loose inequalities will not affect such sensitivities. Now, we implicitly differentiate and obtain:
\definecolor{azuremist}{rgb}{0.94, 1.0, 1.0}
\definecolor{azure}{rgb}{0.0, 0.5, 1.0}
\begin{align*}
& {{dQ}} x + Q  {{dx}} + {{dc}} + {{dA}}^\top \mu + A^\top {{d\mu}} + {{dG}}^\top \lambda + G^\top {{d\lambda}} = 0 \\
& {{dA}} x + A {{dx}} - {{db}} = 0 \\
& {{d\lambda_i}} (h - Gx)_i + \lambda_i ({{dh}} - {{dG}}\ x - G\ {{dx}})_i = 0 \,\,\,\forall i,
\end{align*}
where differentials $dQ$, $dG$, $dh$, $dA$, and $db$ will be treated as input sensitivities or the direction of the desired directional derivative. On the other hand, $dx$, $d\lambda$ and $d\mu$ will be treated as output sensitivities or the directional derivative in the previously-mentioned direction.  Hence, $dQ \in \Reals^{n\times n}$, $dG \in \Reals^{p \times n}$, $dh \in \Reals^p$, $dA \in \Reals^{m\times n}$, $db \in \Reals^m$, $dx \in \Reals^n$, $d\lambda \in \Reals^p$ and $d\mu \in \Reals^m$.

Regrouping the differential forms of parameters and solution variables results in the following system:
\begin{gather*}
\begin{bmatrix}
Q & G^\top & A^\top\\
D(\lambda) G & D(h - Gx) & 0\\
A & 0 & 0
\end{bmatrix}
\begin{bmatrix} dx \\ d\lambda \\ d\mu \end{bmatrix}
=
-
\begin{bmatrix}
dQ\ x + dc + dG^\top\lambda + dA^\top \mu\\
D(\lambda) dh - D(\lambda) dG\ x \\
dA x - db
\end{bmatrix},
\end{gather*}
\noindent
where $D(\cdot)$ denotes the diagonal matrix formed from a given vector.
The system above can be used to compute several quantities of interest.
For given values of the parameters $\theta = (c, Q, A, G, b, h)$ , a primal-dual solution $x, \lambda, \mu$ and sensitivities $dQ$, $dG$, $dh$, $dA$, and $db$,
solving the system can let us derive sensitivities on $dx$, $d\lambda$ and $d\mu$. In other words, we obtain the directional derivatives of $dx$, $d\lambda$ and $d\mu$ with respect to the direction $dQ$, $dG$, $dh$, $dA$, and $db$. By setting the value of one to a single entry among $dQ$, $dG$, $dh$, $dA$, and $db$ and zero to all others, we can obtain a column of the Jacobian matrix. Repeating the process, we can obtain the full Jacobian matrix.
However, in most applications where derivative information is needed, we only
need to perform Jacobian-vector products (JVP) or vector-transpose-Jacobian 
products (VJP). JVP is equivalent to the above-mentioned directional derivatives (if the vector in question is the input sensitivity: $dQ$, $dG$, $dh$, $dA$, and $db$). VJP, on the other hand, is slightly more complicated, but there is a closed-form solution that requires a single linear system solution. We omit the formulas and refer the reader to \cite{amos_optnet_2019}.
Thus, exposing the two principal modes of differentiation
present in differentiable libraries or Automatic Differentiation (AD) tools:
forward- and reverse-mode AD.
%

\subsection{Conic programs}
\label{sec:conic}

Although QPs capture numerous problems of interest, many problems require a richer set of constraints which can be modeled as conic constraints.
Convex conic optimization problems have standard closed-form primal and dual expressions,
\begin{align*}
\textbf{Primal Problem} & & \textbf{Dual Problem} & \\
\min_{x \in \Reals^n} \quad & c^\top x   \quad & \max_{y \in \Reals^m} \quad & - b^\top y  \\
\mbox{s.t.} \quad & A x + s = b  \quad & \mbox{s.t.} \quad & A^\top y + c = 0 \\
& s \in \mathcal{K} &  & y \in \mathcal{K}^*
\end{align*}
where $x$ is the primal variable, $y$ is the dual variable, $s \in \Reals^m$ is the primal slack variable, $\mathcal{K} \subseteq \Reals^m$ is a closed convex cone and $\mathcal{K}^* \subseteq \Reals^m$ is the corresponding dual cone. Note that in above form, $A \in \Reals^{m \times n}$, $b \in \Reals^m$, $c \in \Reals^n$ are problem data.

The methodology of $\partial \mathcal{O}$ has been developed for conic problems in
\citet{amos_differentiable_2019} and \citet{agrawal_differentiating_2020} in parallel
and extended in \citet{agrawal_differentiable_2019}.
The conic case is more involved due to the conic constraints. We follow the notation of \citet{agrawal_differentiating_2020} closely for simpler referencing, but we describe the method in a complete and self-contained fashion.
Instead of deriving the necessary system of equations from the KKT conditions, the proposed approach will rely on the Homogeneous Self-Dual Embedding (HSDE) \citep{scs_solver, busseti_solution_2019}. By construction, a solution of the HSDE will be a solution to the KKT system and the optimization problem in question. The HSDE is written as:
\begin{align*}
& Q u  = v, \quad u = (x, y, \tau)\in \Reals^n \times \mathcal{K} \times \Reals_+, \quad v = (r, s, \kappa)\in \{0\}^n \times \mathcal{K} \times \Reals_+
\end{align*}
Where $r, \tau, \kappa$ are additional variables created to construct the HSDE, and:
\begin{equation*}
    Q = \begin{bmatrix}
         0   & A^\top & c \\
        -A   & 0   & b \\
        -c^\top & -b^\top & 0
    \end{bmatrix}
\end{equation*}
We apply the change of variables $M^{-1}(u, v) = u - v, M(z) = (\Pi z, \Pi z - z)$ from \citet{busseti_solution_2019}, where $\Pi$ is the projection operator onto the Cartesian product $\Reals^n \times \mathcal{K}^* \times \Reals_+$, to obtain a reformulation of the HSDE as a root-finding  problem:
\begin{align*}
& \mathcal{N}(z, Q) = 0, \quad z_{n+m+1} \neq 0
\end{align*}
where:
\begin{equation*}
    \mathcal{N}(z, Q) = ((Q - I) \Pi + I) \left(\frac{z}{|z_{n+m+1}|}\right),
\end{equation*}
is the so-called normalized residual map (NRM), defined in \citet{busseti_solution_2019}.

In this case, the solution map from the problem data to the primal-dual pair, $(x, y, s) = \psi(A,b,c)$, is represented as a composition of three functions: $\psi = \phi \circ \mathcal{S} \circ \mathcal{Q}$. The definition of each of these functions and their derivatives are given by:

\begin{enumerate}
\item $\mathcal{Q}$ maps the problem data, $(A, b, c)$ to the a skew-symmetric matrix $Q \in \Reals^{(n+m+1)\times(n+m+1)}  $:
\begin{equation*}
    Q = \begin{bmatrix}
         0   & A^\top & c \\
        -A   & 0   & b \\
        -c^\top & -b^\top & 0
    \end{bmatrix}, \text{and its differential is: }
    d\mathcal{Q} = \begin{bmatrix}
         0   & dA^\top & dc \\
        -dA   & 0   & db \\
        -dc^\top & -db^\top & 0
    \end{bmatrix}
\end{equation*}
\item $\mathcal{S}$ maps the $Q$ matrix from the HSDE into the zero of the normalized residual map: $z \in \Reals^{n+m+1}$.
The key point is that, given a matrix $Q$, the solution $z$ of $\mathcal{N}(z, Q) = 0$ will lead to the solution of the HSDE and, ultimately, to the primal-dual solution. Hence, it is possible to implicitly differentiate the equation $\mathcal{N}(\mathcal{S}(Q), Q) = 0$ and obtain:
\begin{align*}
    & D\mathcal{S}(Q) = -(D_z \mathcal{N}(\mathcal{S}(Q), Q))^{-1} D_Q \mathcal{N}(\mathcal{S}(Q), Q),
\end{align*}
where
\begin{align*}
    & D_Q \mathcal{N}(z, Q)\left[U\right] = U\Pi\left(\frac{z}{z_{n+m+1}}\right)\\
    & D_z \mathcal{N}(z, Q) =  \frac{((Q - I) D\Pi(z) + I)}{z_{n+m+1}} - \sign(z_{n+m+1})((Q - I)\Pi + I)\left(\frac{z}{z_{n+m+1}^2}\right) e_{n+m+1}^\top\\
\end{align*}
where $D_x f(x,y)[w]$ denotes the directional derivative of $f$ at $x$ in direction $w$ and
$e_{n+m+1}$ the basis vector with 1 at index $n+m+1$.
Note that the second term of $D_z \mathcal{N}(z, Q)$ vanishes if $z$ is a solution of the HSDE.
\item $\phi$ maps the solution of the HSDE to the primal-dual pair:
\begin{equation*}
(x, y, s) = \phi(z) = (z_{1:n}, \Pi_{\mathcal{K}^{*}}(z_{n+1:n+m}), \Pi_{\mathcal{K}^{*}}(z_{n+1:n+m}) - z_{n+1:n+m})/z_{n+m+1}.
\end{equation*}
Its derivative is given by:
\begin{equation*}
    D\phi(z) = \begin{bmatrix}
        I & 0 & -x \\
        0 & D\Pi_{\mathcal{K}^{*}}(z_{n+1:n+m})  & -y \\
        0 & D\Pi_{\mathcal{K}^{*}}(z_{n+1:n+m}) -I & s
    \end{bmatrix}
\end{equation*}
where $\Pi_{\mathcal{K}^{*}}$ is the projection onto $ \mathcal{K}^*$.
The derivatives of projections onto classic cones like the positive orthant, second-order, positive semidefinite, and exponential cones are given in \citet{busseti_solution_2019} and implemented for DiffOpt in \texttt{MathOptSetDistances.jl}~\citep{mathieu_besancon_2022_6505452}.
Exponential cone projections are performed using
the technique from \citet{friberg2021projection}.
\end{enumerate}
Finally, we obtain:
\begin{equation*}
(dx, dy, ds) = D\psi(A, b, c) (dA, db, dc) = D\phi(z) D\mathcal{S}(Q) D Q (A, b, c) (dA, db, dc),
\end{equation*}
which can be used directly to compute JVP (directional derivatives) from the sensitivities $(dA, db, dc)$ by performing the above-described computations. Again, VJP is slightly more complex but also has closed-form solutions that require a single linear system solution. For more details, see \cite{agrawal_differentiating_2020}.

\section{Package structure}\label{sec:structure}

DiffOpt is a Julia package that offers differentiable optimization algorithms to the JuMP ecosystem. In order to integrate seamlessly with other packages, DiffOpt is built on top of \texttt{MathOptInterface.jl} (MOI), a foundational unifying package
for constrained optimization, designed to be a backend for modeling
interfaces such as \texttt{JuMP.jl} or \texttt{Convex.jl}.
MOI allows the user to describe structured optimization problems in a unified format based on the constraint representation of functions of variables belonging to sets.
This abstraction covers a wide variety of problems, including linear, quadratic, and conic constraints, as well as 
more specific sets, such as Special Ordered Sets or complementarity constraints.
Moreover, MOI includes a \textit{bridging} mechanism turning the
user-provided problem into a problem structure that the chosen solver accepts through successive transformations (or bridges) of the function-set pairs.
As detailed in \citet[Section~2.1.2]{legat2020set},
the constraint types and bridges form a \emph{directed hypergraph} where each function-set pair is a node and each bridge is a \emph{single source hyperedge}.
The source of the hyperedge is the constraint type that is transformed by the bridge and the targets of the hyperedge are the set of constraints that are created by the transformation.
Illustrations of such hypergraph can be found in \citet[Figure~2.2, 2.3]{legat2020set}.
Given a constraint of the user model,
the choice of transformation to apply can
then be reformulated as the search for a \emph{hyperpath} starting at the corresponding node and ending at nodes corresponding to constraint types supported by the solver.

The design of MOI enables extending its interface through various mechanisms.
First, constraints are defined as pairs \texttt{func}-in-\texttt{set} where \texttt{func} is a function of the decision variables
and \texttt{set} is a set in which the value of the function should belong when evaluated at feasible points.
Solvers that can exploit special problem structure can therefore allow the user to communicate it by defining new function or set types.

Second, most of the MOI interface is built on top of \emph{attributes}.
This enables both the user to communicate custom information to the solver, such as starting values or callbacks, but also the solver to communicate custom results, such as basis status for simplex solvers or Irreducible Inconsistent Subsystems (IIS) for infeasible instances.

Third, MOI optimizers support a layered structure to combine several features.
These layers also support custom constraints and attributes that are typically defined by an inner layer thanks to the well-posed API of MOI.
These layers are commonly referred to as \emph{meta-solvers} as they form solvers parameterized by other solvers.
Bridges are defined as MOI layers that transform constraints into constraints of different types for their inner layer.
When defining a new constraint type, defining bridges to transform it into classical constraint types allows the user
to encode this special structure in models while still being able to use solvers not supporting this structure.
Below are a few examples of MOI extensions that illustrate this.

Some or all of the above features of MOI were previously used in extensions and motivated the design of DiffOpt in its current form. We highlight some:
\begin{enumerate}
    \item
\texttt{SumOfSquares.jl} defines the Sum-of-Squares cone as a new set type~\citep{weisser2019polynomial}
and \emph{Gram matrices}, \emph{Moment matrices} and \emph{Sum-of-Squares decompositions} as new attributes.
It then defines a bridge for transforming Sum-of-Squares constraints into semidefinite constraints.
This bridge is then automatically applied only if the solver does not support Sum-of-Squares constraints but does support semidefinite constraints.
For instance, it will be applied for classical semidefinite programming solvers for all semidefinite programming solvers with the exception of
\texttt{Hypatia.jl}~\citep{coey2020solving} which supports Sum-of-Squares constraints natively.

    \item 
\texttt{Dualization.jl} \citep{guilherme_bodin_2021_4718987}
offers a dualization meta-solver. The optimization problem is automatically converted to its dual form and reaches the internal solver only in dual form, of which the solution is then mapped back to the user.

\item
\texttt{QuadraticToBinary.jl}~\citep{joaquim_dias_garcia_2021_4718981} is another meta-solver that converts quadratically-constrained problems into Mixed Integer Linear Programs by automatically applying binary expansions.

\item
\texttt{ConstraintSolver.jl}~\citep{constraintsolver} is a  constraint programming solver that has defined new sets and functions following the MOI standard.

\end{enumerate}

DiffOpt can be viewed as five main components and aspects that are covered in the following sections.
First, it extends MOI by creating new attributes allowing the user and solver to communicate forward or reverse differentiation input and output (Section \ref{sec:interface}).
Second, it implements the quadratic and conic problem differentiation rules described in \cref{sec:background} as MOI models implementing these attributes (Section \ref{sec:rules}).
Third, it implements the communication of these attributes through the MOI caching and bridging layers (Sections \ref{sec:bridges} and \ref{sec:quad_to_soc}).
Fourth, it implements a meta-solver that implements the computation of the differentiation attributes and is parameterized by a solver that should support solving the problem (Sections \ref{sec:meta_solver} and \ref{sec:diff_solver}).
Fifth, it allows to integrate optimization layers into AD systems in Julia using the \texttt{ChainRulesCore.jl} package (Section \ref{sec:chain}). 

\subsection{Interface}
\label{sec:interface}
After the problem is solved, the user can pass parameter perturbations in case of forward differentiation or sensitivities with respect to solution variable values in the case of reverse differentiation (or even both).
These are passed to the \texttt{DiffOpt.Optimizer} using the attributes detailed in \cref{tab:attr:in}.
After sensitivities are loaded, the user might call \texttt{DiffOpt.forward\_differentiate!} or \texttt{DiffOpt.reverse\_differentiate!} to compute the derivatives with respect to the input sensitivities. The resulting derivatives are queried again as typical solver attributes, detailed in \cref{tab:attr:out}.

\begin{table}[!ht]
\centering
\begin{small}
\begin{tabular}{|l|l|}
\hline
\texttt{ForwardObjectiveFunction} & Forward-mode tangent for the objective function \\
\texttt{ForwardConstraintFunction} & Forward-mode tangent for a constraint function \\
\texttt{ReverseVariablePrimal} & Reverse-mode tangent for a variable value \\
\hline
\end{tabular}
\end{small}
\caption{Differential Optimization attributes (DiffOpt attributes for short) for passing perturbations and sensitivities.}
\label{tab:attr:in}
\end{table}

\begin{table}[!ht]
\centering
\begin{small}
\begin{tabular}{|l|l|}
\hline
\texttt{ForwardVariablePrimal} & Forward-mode tangent for a variable value \\
\texttt{ReverseObjectiveFunction} & Reverse-mode tangent for the objective function \\
\texttt{ReverseConstraintFunction} & Reverse-mode tangent for a constraint function \\
\hline
\end{tabular}
\end{small}
\caption{Differentiable optimization attributes (DiffOpt attributes for short) for querying resulting derivatives.}
\label{tab:attr:out}
\end{table}
Perturbations are passed and queried in the form of MOI functions (affine functions and quadratic functions) associated with constraints or the objective. The coefficients of variables (or quadratic terms) are the perturbations related to those variables (or quadratic terms) and the respective associated constraint or objective. This API allows the perturbation to flow through model transformations defined by the above-mentioned \textit{bridges}. To keep the API efficient, the returned functions are lazily computed so that only the coefficients required by the final user are actually evaluated.

DiffOpt integrates smoothly with \texttt{JuMP} just like any solver implementing the MOI interface. Because derivative-related attributes are slightly more complex than traditional attributes, JuMP was added as a dependency so that we could overload a small subset of the JuMP API to achieve a better user experience. 
Moreover, DiffOpt can communicate with any solver that has an MOI interface, more than 40 of them are listed in the JuMP Manual \citep{jumpmanual}.
The only computationally demanding step other than solving the optimization problem is the linear system solution.
Since the special structure of the constraint and Hessian matrices can be exploited to accelerate the linear system solve, the \textit{DiffOpt model}
exposes the attribute \texttt{LinearAlgebraSolver} that can be set to implement a specialized solving method for the linear system resulting from the constraints,
allowing users to swap and try different methods and linear algebra solvers.
This generic method to handle and solve the linear system was inspired by \texttt{LinearSolve.jl} \citep{linsolve} with a simplified interface.

\subsection{Differentiation rules as MOI models}
\label{sec:rules}
DiffOpt implements the differentiation rules for QPs (resp. CPs)
described in \cref{sec:background} as a \emph{differentiable optimization MOI model} (\textit{DiffOpt model} for short) \path{DiffOpt.QuadraticProgram.Model} (resp. \path{DiffOpt.ConicProgram.Model}).
These \textit{DiffOpt models} represent QPs (resp. CPs) in the matrix standard form described in \cref{sec:background}.
They do not support solving the QP (resp. CP) but they support having the primal and dual solution being set by the user.
Then, they support the differentiation API described in \cref{sec:interface}.

The QP and CP standard forms described in \cref{sec:background} are, however, strict on the constraints accepted for optimization models.
For instance, the QP form supports inequality constraints $\smash{Gx \le h}$, i.e., \texttt{MOI.ScalarAffineFunction}-in-\texttt{MOI.LessThan}, but not $\smash{Gx\ge h}$, i.e., \texttt{MOI.ScalarAffineFunction}-in-\texttt{MOI.GreaterThan}.
This requires the user to transform their model in order to fit the solver-compatible representation.
It is then tedious and error-prone to map both the primal and dual solutions through these transformations as well as the DiffOpt attributes.
Fortunately, as described in the next section, the transformation of these attributes is implemented through bridges.

Therefore, adding a bridging outer layer on top of \texttt{DiffOpt.QuadraticProgram.Model} or \texttt{DiffOpt.ConicProgram.Model} allows the user to
model the QP or CP in the most convenient form while all these transformations are carried out transparently.
Moreover, it also allows extending DiffOpt to new problem classes through bridges.
For instance, as \texttt{SumOfSquares.jl} defines the transformation from a Sum-of-Squares constraint to a semidefinite constraint using a bridge, defining how to transform the DiffOpt attributes through these bridges automatically broadens the class of programs supported by
\texttt{DiffOpt.ConicProgram.Model} to Sum-of-Squares programs. Another example of the proposed design generality is a possible integer programming differentiation methodology, in this case, the basic foundations of DiffOpt would be ready, but a new {DiffOpt model} named \texttt{IntegerDiffProblem}  would be required, analogous to \texttt{DiffOpt.QuadraticProgram.Model} and \texttt{DiffOpt.ConicProgram.Model}.

\subsection{Affine model transformations}\label{sec:bridges}
Most model transformations rely on an affine relation between two sets $\mathcal{S}_1 \subseteq \mathbb{R}^n$, $\mathcal{S}_2$ of the form
\begin{equation}
  \label{eq:s12}
  \mathcal{S}_1 = \{\, x \in \mathbb{R}^n \mid
  \exists u \in \mathbb{R}^m \text{ s.t. }
  Ax + Bu + c \in \mathcal{S}_2 \,\}.
\end{equation}

As detailed in \citet[Section~2.1.2]{legat2020set},
there is an automated way to implement the transformation for primal and dual results
given the transformation data $A$, $B$ and $c$.
Similarly, we develop in this section the transformation for the DiffOpt attributes.

Consider the extension of MOI to new sets $\mathcal{S}_1$ and $\mathcal{S}_2$ satisfying \cref{eq:s12}.
The two sets are created as \texttt{MOI.AbstractSet}s \texttt{S1} and \texttt{S2}.
The scalar products $\langle x, y \rangle_i$
between a vector $x$ of $\mathcal{S}_i$ and a vector $y$ of its dual are defined
by implementing a new method to \texttt{MOI.Utilities.set\_dot(x, y, Si())}.
By default, if no method \texttt{MOI.Utilities.set\_dot} is defined for \texttt{Si}, the inner product falls back to ${x^\top y}$.
These scalar products can be arbitrary since \texttt{MathOptInterface}, \texttt{Dualization} and \texttt{DiffOpt} never assume it to be $x^\top y$ and always use the generic \texttt{set\_dot} function.
A different scalar product is, for instance, defined for the set \texttt{MOI.PositiveSemidefiniteConeTriangle}
so that the scalar product between the vectorization upper triangular matrices corresponds to the Frobenius inner product between them.

The transformation allowed by \cref{eq:s12} is implemented as a bridge that,
given a constraint $f_1(x) \in \mathcal{S}_1$,
creates variables $u \in \mathbb{R}^m$
and transforms the constraint
into a constraint $f_2(x, u) \in \mathcal{S}_2$
where $f_2(x, u) = Af_1(x) + Bu + c$.

We have $\dd{f_2} = A\dd{f_1}$. Therefore, we see that the \texttt{ForwardConstraintFunction}
forward-mode tangent $\Delta f_1$ should be mapped to $A \Delta f_1$ and the
\texttt{ReverseConstraintFunction} reverse-mode tangent $\Delta f_2$
should be mapped to $A^* \Delta f_2$, with $A^*$
the adjoint matrix of $A$ satisfying
\[ \langle Ax, y \rangle_2 = \langle x, A^* y \rangle_1 \]
for all $x$ in the space of $\mathcal{S}_1$ and $y$ in the dual space of $\mathcal{S}_2$.

\subsection{Quadratic and second-order cone model transformations}\label{sec:quad_to_soc}
At the time of writing, the only bridge in MOI not based on such an affine relation of the form~\eqref{eq:s12} is the transformation from a convex quadratic constraint:
$$\frac{1}{2}x^\top Q x + a^\top x + \beta \le 0,$$
to the affine conic constraint:
$$(1, -a^\top x - \beta, Ux) \in \mathcal{R},$$
where $\mathcal{R}$ is the rotated second order cone of appropriate dimension and $U$ is obtained from the Cholesky decomposition $Q = U^\top U$. Indeed, the Cholesky decomposition is not a linear map from $Q$ to $U$.

The transformation of an optimal dual vector for the conic constraints
into the dual of the quadratic constraint is quite different from bridges based on affine relations discussed in \cref{sec:bridges}.
In MOI, its implementation relies on the complementary slackness and
a geometric property of the second-order cone~\cite[Example~2.1.2]{legat2020set}.
Transforming DiffOpt attributes requires solving Lyapunov-like matrix equations we detail below.

The ability to transform DiffOpt attributes through this bridge
automatically enables convex quadratically-constrained quadratic programs (QCQPs) to be differentiated, even though they do not correspond to either of the original forms of differentiable models.
Indeed, these can be transformed into a conic form and then differentiated using the conic \textit{DiffOpt model}.
Moreover, quadratic objectives and quadratic constraints can also be mixed with arbitrary conic constraints as well.
While the user could have reformulated these quadratic constraints and quadratic objective into the conic form manually,
they would have obtained the sensitivities in the conic form.
With the transformation presented in this section, the user does not have to perform this reformulation manually and, more importantly, can obtain the sensitivities in the quadratic form.

In the reformulation, the only part that is not an affine transformation (which can be dealt with \cref{sec:bridges}) is the Cholesky decomposition $Q = U^\top U$.
We show in \cref{sec:quadtosoc_forward} (resp. \cref{sec:quadtosoc_reverse}) how to differentiate through this decomposition in forward-mode (resp. reverse-mode).

\subsubsection{Forward-mode}
\label{sec:quadtosoc_forward}

\noindent
For forward-mode differentiation,
given a symmetric matrix $\dd{Q}$, we want to find the corresponding tangent matrix $\dd{U}$ for the Cholesky factor.
Differentiating the Cholesky decomposition relation $Q = U^\top U$, we obtain:
\begin{equation}
  \label{eq:dQdU}
  \dd{Q} = \dd{U}^\top U + U^\top \dd{U}.
\end{equation}
\Cref{eq:dQdU} has a Lyapunov-like form that has been extensively studied in \cite{de2011consistency,djordjevic2007explicit,braden1998equations}.
Typical models tackled in optimization are large and sparse, the positive definite matrix $Q$ is therefore represented in a sparse data structure in MOI; the Cholesky factor $U$ is obtained with the SuiteSparse CHOLMOD library~\citep{davis2019algorithm}.
The factor $U$ can then be assumed to be invertible and have an upper triangular structure.
The unique upper triangular solution $\dd{U}$ can therefore be obtained using
\cref{prop:dU_form_dQ} below.
As the right-hand side of \cref{eq:dU_from_dQ}
only depends on the $k-1$ first columns of $\dd{U}$,
this allows computing each column of $\dd{U}$, in increasing column index order,
by solving the triangular linear system \cref{eq:dU_from_dQ}.
Let $[k]$ denote the set $\{1, \ldots, k\}$.

\begin{proposition}
  \label{prop:dU_form_dQ}
  Given a symmetric matrix $B \in \mathbb{R}^{n \times n}$ and a matrix $A \in \mathbb{R}^{n \times n}$ such that $A_{[k],[k]}$ is invertible for all $k \in [n]$,
  the equation
  \begin{align}
    \label{eq:BXA}
      X^\top A + A^\top X = B.
  \end{align}
  has a unique upper triangular solution $X$.
  This solution satisfies
  \begin{equation}
      \label{eq:dU_from_dQ}
      A_{[k],[k]}^\top X_{[k],k}
      = \begin{bmatrix}
          B_{[k-1],k} - X_{[k-1],[k-1]}^\top A_{[k-1],k}\\
          B_{k,k}/2
      \end{bmatrix}.
  \end{equation}
  \begin{proof}
    As $B$ is symmetric,
    \cref{eq:BXA} only has to be verified
    for $B_{[k],k}$ for all $k = 1, \ldots, n$.
    For any $k = 1, \ldots, n$, we have
    $$
     X_{[n],[k]}^\top A_{[n], k} + A_{[n], [k]}^\top X_{[n],k} = B_{[k],k}
    $$
    So
    $$A_{[n], [k-1]}^\top X_{[n],k} = B_{[k-1],k} - X_{[n],[k-1]}^\top A_{[n], k}$$
    and
    $$A_{[n], k}^\top X_{[n],k} = B_{k,k}/2.$$
    Since $X$ is upper triangular,
    this is equivalent to \cref{eq:dU_from_dQ}.
    As $A_{[k],[k]}$ is invertible, \cref{eq:dU_from_dQ} holding for all $k$ ensures the existence and unicity of the solution.
  \end{proof}
\end{proposition}

\subsubsection{Reverse-mode}
\label{sec:quadtosoc_reverse}

For reverse-mode differentiation,
given a matrix $\Delta U$, we want to find a symmetric matrix $\Delta Q$ such that,
for any symmetric matrix $\dd{Q}$, we have
\begin{equation}
  \label{eq:DQDU}
  \langle \dd{Q}, \Delta Q \rangle = \langle \dd{U}, \Delta U \rangle.
\end{equation}
where $\dd{U}$ is the unique lower triangular solution of \cref{eq:dQdU} and $\langle \cdot, \cdot \rangle$ is the scalar product $\langle A, B \rangle = \tr(A^\top B)$.
By \cref{eq:dQdU}, \cref{eq:DQDU} is equivalent to
\begin{equation}
  \label{eq:DQDU2}
  \langle \dd{U}^\top U + U^\top \dd{U}, \Delta Q \rangle = \langle \dd{U}, \Delta U \rangle.
\end{equation}
As $\Delta Q$ is symmetric, \cref{eq:DQDU2} is equivalent to
\begin{align}
  \notag
  2\langle U^\top \dd{U}, \Delta Q \rangle
  & = \langle \dd{U}, \Delta U \rangle\\
  \label{eq:DQDU3}
  2\langle \dd{U}, U \Delta Q \rangle
  & = \langle \dd{U}, \Delta U \rangle.
\end{align}
Let $\triu(\cdot)$ denote the upper-triangular part of a matrix.
\Cref{eq:DQDU3} is satisfied for all upper triangular $\dd{U}$ if and only if
$2 \triu(U \Delta Q) = \triu(\Delta U)$.
Since $U$ is invertible and upper triangular (as discussed in \cref{sec:quadtosoc_forward}),
\cref{prop:DQ_form_DU} allows computing each column of $\Delta Q$, in decreasing column index order, by solving the triangular linear systems \eqref{eq:DQ_from_DU}.

\begin{proposition}\label{prop:DQ_form_DU}
  Given a matrix $B \in \mathbb{S}^{n}$ and an invertible upper triangular matrix $A \in \mathbb{R}^{n \times n}$ such that $A_{[k],[k]}$ is invertible for all $k \in [n]$, the equation
  \begin{align}
    \label{eq:AXB}
      \triu(AX) = \triu(B)
  \end{align}
  has a unique symmetric solution $X$. This solution satisfies
  \begin{equation}
      \label{eq:DQ_from_DU}
      A_{[k],[k]} X_{[k],k} = B_{[k],k} - \sum_{i=k+1}^n A_{[k],i} X_{k, i}.
  \end{equation}
  for all $k \in [n]$
  \begin{proof}
    As the equation is on the upper-triangular part of the matrix,
    \cref{eq:BXA} only has to be verified for $B_{[k],k}$ for all $k = 1, \ldots, n$.
    For any $k = 1, \ldots, n$, we have
    $$
    A_{[k],[n]}^\top X_{[n],k} = B_{[k],k}
    $$
    hence
    $$
    A_{[k],[k]}^\top X_{[k],k} = B_{[k],k}
    - \sum_{i=k+1}^n A_{[k],i} X_{i, k}.
    $$
    Since $X$ is symmetric,
    $X_{i, k}$ is equal to $X_{k,i}$ so this is equivalent to \eqref{eq:DQ_from_DU}.
    Since $A_{[k],[k]}$ is invertible, \cref{eq:DQ_from_DU} holding for all $k$ ensures the existence and unicity of the solution.
  \end{proof}
\end{proposition}

\subsection{Meta-solver}
\label{sec:meta_solver}
DiffOpt is designed as a meta-solver with a structure illustrated in \cref{fig:design}. The main structure made available by the package is \texttt{DiffOpt.Optimizer} that is parameterized by a mathematical programming inner solver.
This inner solver may be any object supporting storing and solving the problem provided by the user.
As shown in \cref{fig:design}, the entry point is a cache that is added to ensure efficient storage and access to the user model. Also, bridge layers are added as the user model may use constraint types that are not natively supported by the inner solver (or the \textit{DiffOpt models}) and need to be transformed.
The solution found by the inner solver is communicated to the \textit{DiffOpt model} that caches matrix form data as described in \cref{sec:rules}.
This \textit{DiffOpt model} is then used to compute forward and/or reverse differentiation tangents.

Note that the bridges used for the inner solver and for the \textit{DiffOpt model} are completely independent. They can be entirely different as the primal and dual results are automatically transformed through the inner solver bridge layer from the solver solution into the solution corresponding to the user model and then through the \textit{DiffOpt model} bridges into the solution corresponding to the \textit{DiffOpt model} standard form.

This design of \path{DiffOpt.Optimizer} also enables adding new \textit{DiffOpt models} in addition to \path{DiffOpt.QuadraticProgram.Model} and \path{DiffOpt.ConicProgram.Model} in order to further broaden the class of supported models.

\begin{figure}[!ht]
\begin{footnotesize}
    \begin{center}
\Tree
[
    [
        [
            {Inner solver}
        ].Bridges
        [
            \texttt{QuadraticProgram.Model}
            \texttt{ConicProgram.Model}
        ].Bridges
    ].Cache
].\texttt{DiffOpt.Optimizer}
    \end{center}
\end{footnotesize}
    \caption{Design of the \texttt{DiffOpt.Optimizer} structure}
    \label{fig:design}
\end{figure}

Because MOI does not distinguish between the two considered classes, QP and CP, the user can pass any of the two problem classes to DiffOpt without ever having to know which of the two methods will be used. DiffOpt automatically selects the appropriate problem class based on the type of constraints and objective function.

In the following code excerpt, we demonstrate the usage of DiffOpt, starting from a simple JuMP model, then going through a reverse differentiation procedure.

\begin{minipage}{\linewidth}
\begin{jllisting}
using JuMP, DiffOpt, Clp

model = JuMP.Model(() -> diff_optimizer(Clp.Optimizer))
@variable(model, x)
@constraint(model, cons, x >= 3)
@objective(
    model, 
    Min, 
    2x,
)

optimize!(model) # solve

MOI.set.(  # set perturbations / gradient inputs
    model, 
    DiffOpt.ReverseVariablePrimal(), 
    x, 
    1.0,
)
DiffOpt.reverse_differentiate!(model) # differentiate

# fetch expression of the gradient of constraint
grad_exp = MOI.get(   # -3x+1
    model, 
    DiffOpt.ReverseConstraintFunction(), 
    cons
)
JuMP.constant(grad_exp)  # 1
JuMP.coefficient(grad_exp, x)  # -3
\end{jllisting}
\end{minipage}

\subsection{Differentiable solvers}\label{sec:diff_solver}

Solvers may have information that would help in computing derivative information that is not part of the primal and dual results.
For instance, they may have computed an LDLT factorization that could speed up the differentiation of the problem.
As a matter of fact, OSQP~\citep{osqp} is implementing differentiation with respect to problem data by reusing parts of the computation carried out when solving the problem.
Another example is sIpopt~\citep{pirnay2012optimal} which reuses matrix factorizations computed by Ipopt.
As the DiffOpt interface is built on MOI attributes, it can naturally be implemented on the solver side.

\subsection{Building differentiable pipelines with ChainRules primitives}\label{sec:chain}

Automatic Differentiation (AD) has become a cornerstone of
machine learning, evolving from the static transformation of program sources to a fully dynamic process \citep{innes2019differentiable}.
The Julia AD landscape has been evolving rapidly \citep{schafer2021abstractdifferentiation}
and is now converging toward a shared set
of derivative primitives implemented for elementary functions.
These elementary derivatives are then exploited by AD libraries to compute derivatives of full programs by the chain rule.
The primitives are defined with a common interface in the \texttt{ChainRules.jl} library \citep{chainrules}, including rule definitions for forward-, reverse- or mixed-mode AD. In addition to defining a grammar to declare rules for elementary functions, it implements these rules for functions from Julia \texttt{Base} and standard libraries.
Specifically, the methods that have to be implemented to provide derivative information are
defined in \texttt{ChainRulesCore.jl}~\citep{chainrules} and implemented in
\texttt{ChainRules.jl} for Julia \texttt{Base} and standard
library functions.

The \texttt{ChainRules.jl} system reasons
on the differentiation of a function's output with respect to its 
inputs while the DiffOpt interface is based on MOI and, thus, on
the incremental construction of a model represented as a
single mutable object.
One approach is to construct an implementation of the solution map, which takes as an argument the model parameters, builds and optimizes the model object, and returns the optimal solution.
The solution map is a pure function and its derivatives can 
be expressed in terms of \texttt{ChainRules.jl} primitives and
implemented using derivative information from DiffOpt.

This implementation allows external users to effortlessly bring
any MOI model built directly or through
a modelling interface like \texttt{JuMP} or \texttt{Convex.jl} to a differentiable pipeline,
regardless of the underlying solver used to produce the optimal solution.
However, the user still needs to write down the solution map
and implement the \texttt{ChainRules.jl} interface functions \texttt{frule} or \texttt{rrule} for forward
or reverse differentiation, respectively.
We will provide an example of building such a differentiable pipeline in \cref{sec:convexneuralnet}.

\section{Application examples}\label{sec:examples}

We discuss how differentiating an optimization program allows a variety of applications for different computational tasks.
These examples use various convex solvers to compute the primal and dual solutions in order to highlight the
ease to swap a solver for another in a single line.

All the following examples can be seen in detail in the \textit{Tutorials} section of the DiffOpt manual. The code and data use version 0.4 of the package.

\subsection{Sensitivity Analysis}

Sensitivity analysis 
\citep{saltelli2004sensitivity,bonnans2013perturbation,fiacco1983introduction,robinson1982generalized} focuses on studying how the changes in the inputs of a
mathematical model affects its output.
Sensitivities of a JuMP model can be computed automatically using the previously described methods.
We illustrate sensitivity analysis for a classification and a regression task.

\subsubsection{Classification using SVM}
\label{sec:svm}

Support vector machines (SVM) classify labeled data points with the hyperplane minimizing the norm of classification errors on all points (or achieving the largest margin if the two classes are separable).
Assuming $X \in \mathbb{R}^{n\times d}$ the feature matrix of $n$ data points with $d$ features and $y$ their labels,
a soft-margin $\ell_1$-SVM with $\ell_2$ regularization can be modelled as:
\begin{align*}
\min_{\xi, w, b}\, & \sum_{i=1}^{n} \xi_i + \lambda \|w\|^2 \\
\text{s.t. }\, & y_i (w^\top X_i + b) \geq 1 - \xi_i \,\,\, \forall i \in 1..n\\
& \xi_i \geq 0 \,\,\, \forall i \in 1..n,
\end{align*}
\noindent
where $e_i$ is the soft margin loss on the $i-th$ data point, $w^\top x = b$ is the SVM hyperplane, $\lambda$ is the regularization parameter.

The plots and code transcripts are from \textit{Sensitivity Analysis of SVM} tutorial.
The model is implemented below and solved using Ipopt \citep{wachter2006implementation}.

\begin{minipage}{\linewidth}
\begin{jllisting}
# N, D, X, y are given
λ = 0.05
model = Model(() -> DiffOpt.diff_optimizer(Ipopt.Optimizer))

# Add the variables
@variable(model, ξ[1:N] >= 0)
@variable(model, w[1:D])
@variable(model, b)

# Add the constraints.
@constraint(model,
    con[i in 1:N],
    y[i] * (dot(X[i,:], w) + b) >= 1 - ξ[i]
)

# Define the objective and solve
@objective(model, Min, λ * dot(w, w) + sum(ξ))
optimize!(model)
\end{jllisting}
\end{minipage}

\begin{figure}
    \centering
    \includegraphics[width=100mm]{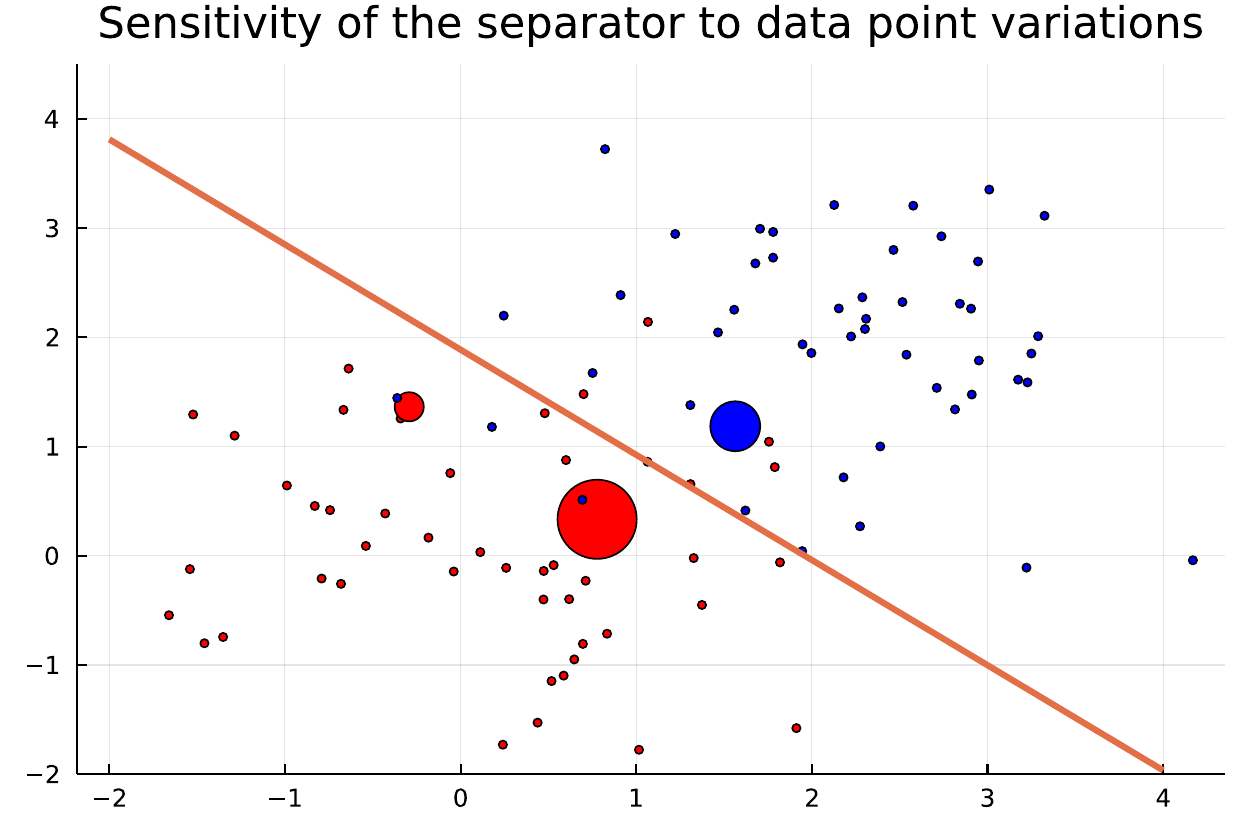}
    \caption{Learned SVM and sensitivities. Data points colors indicate the class and marker size denotes the sensitivity of the hyperplane to $x_i$.}
    \label{fig:senssvm}
\end{figure}

\noindent
Using the forward differentiation mode of DiffOpt, we compute the partial Jacobian with respect to each $i$-th individual
data point and use its norm as the point size in \cref{fig:senssvm}:
\begin{equation*}
\left\| \frac{\partial w}{\partial X_i} \right\|_{2} + \left|\frac{\partial b}{\partial X_i}\right|_{2}.
\end{equation*}
\noindent
The computation is performed as follows:

\begin{minipage}{\linewidth}
\begin{jllisting}
MOI.set(model, DiffOpt.ModelConstructor(), DiffOpt.QuadraticProgram.Model)
∇ = zeros(N)
for i in 1:N
    for j in 1:N
        if i == j
            # identical perturbations on all x_i
            MOI.set(
                model,
                DiffOpt.ForwardConstraintFunction(),
                con[j],
                y[j] * sum(w),
            )
        else
            MOI.set(
                model,
                DiffOpt.ForwardConstraintFunction(),
                con[j],
                0.0,
            )
        end
    end
    DiffOpt.forward_differentiate!(model)
    dw = MOI.get.(
        model,
        DiffOpt.ForwardVariablePrimal(),
        w,
    )
    db = MOI.get(
        model,
        DiffOpt.ForwardVariablePrimal(),
        b,
    )
    ∇[i] = norm(dw) + norm(db)
end
\end{jllisting}
\end{minipage}

A perturbation of the feature matrix $X$ affects the solution and can induce a change in the separating hyperplane decisions $(w, b)$.
Unlike other classification models, not all points affect the hyperplane with small enough perturbations,
the optimal solution depends only on a few data points, the support vectors that name the method.
The impact of these perturbations is displayed in \cref{fig:senssvm}.

Note that, thanks to \cref{sec:quad_to_soc},
the sensitivities can be obtained by the quadratic DiffOpt model but also using conic DiffOpt models.
Indeed, in view of \cref{sec:meta_solver},
even if the model is solved by Ipopt in QP form, the dual of the constraint provided by Ipopt can be transformed into the dual of the rotated second-order cone constraint through the bridge so that the conic DiffOpt model has the dual information needed to compute the DiffOpt attributes.
The DiffOpt attribute in conic form can then be transformed into quadratic form using the results developed in \cref{sec:quad_to_soc}.
That transparently decouples the models used by the differentiation, the solver and the user.
The model used depends on whether \texttt{DiffOpt.ModelConstructor} was set to \texttt{DiffOpt.QuadraticProgram.Model} or
\texttt{DiffOpt.ConicProgram.Model} in the first line.

\subsubsection{Ridge regression sensitivity}

Ridge regression avoids overfitting with an $\ell_2$-norm penalty added to a linear regression model
and is particularly advantageous when the number of features is as large as the number of observations.
Assume $X = \{(x, y)\} \subset \mathbb{R}^{d+1}$ to be the set of $n$ data points.
Then a ridge regression fitting problem can be modeled as:
\begin{align*}
\min_{w,b} \quad & \sum_{i=1}^{N} (y_{i} - w^\top x_{i} - b)^2 + \alpha \|w\|_2^2
\end{align*}
with $\alpha$ the regularization constant.

The plots and code transcripts are from \textit{Sensitivity Analysis of Ridge Regression} tutorial.
We implement and solve below a univariate example with
DiffOpt using Ipopt as the underlying QP solver.

\begin{minipage}{\linewidth}
\begin{jllisting}
# X, Y, N are given

model = Model(() -> diff_optimizer(SCS.Optimizer))

@variable(model, w)
@variable(model, b)
alpha = 0.8  # regularization constant
@expression(model, e[i=1:N], Y[i] - w * X[i] - b)

@objective(
    model,
    Min,
    1 / N * dot(e, e) + alpha * (w^2),
)

optimize!(model)
\end{jllisting}
\end{minipage}

Similar to the SVM example, a change in a single independent or dependent variable value
$x_i$ or $y_i$ for a given data point can affect the learned model.
We use DiffOpt in forward mode to quantify these sensitivities to individual data points:
\begin{equation*}
    \frac{\partial w}{\partial x_i} \text{ and } \frac{\partial w}{\partial y_i}.
\end{equation*}

\noindent
A \texttt{ForwardObjectiveFunction} attribute can be set for the perturbation of the objective function.
It takes as input the expression proportionally dependent on the perturbed parameter $\theta$.
Given a generic expression $f(x; \theta) = \theta g(x)$ with parameter $\theta$, the expected input is
$g(x)$.
A particular aspect here is that the $x_i$ and $y_i$ values appear as linear and quadratic terms in the loss function. If the parameter $\theta$ appears both linearly and quadratically, the corresponding objective perturbation can be derived with a first-order Taylor expansion:
\begin{align*}
&&  f(x; \theta) &= \theta g(x) + \theta^2 h(x) \\
&&  f(x; \theta + \delta) &= (\theta + \delta) g(x) + \theta^2 h(x) + 2 \delta \theta h(x) + \delta^2 h(x) \\
& & f(x; \theta + \delta) &\approx f(x; \theta) + \delta (g(x) + 2\theta h(x)).
\end{align*}
\noindent
When applied to the loss function, the perturbation $\delta^{(x)}_i$, $\delta^{(y)}_i$ on $x_i$, $y_i$ respectively results in first-order perturbation approximations:
\begin{align*}
& \delta^{(x)}_i (2w^2  x_i + 2b  w - 2 w y_i),\\
& \delta^{(y)}_i (2 y_i - 2b - 2w x_i)\\
\end{align*}
\noindent
respectively. Using these input perturbations, we can extract the output sensitivity of the slope $w$
using the \texttt{DiffOpt.ForwardVariablePrimal} variable attribute:

\begin{minipage}{\linewidth}
\begin{jllisting}
MOI.set(model, DiffOpt.ModelConstructor(), DiffOpt.QuadraticProgram.Model)
∇y = zero(X)
∇x = zero(X)
for i in 1:N
    MOI.set(
        model,
        DiffOpt.ForwardObjectiveFunction(),
        2w^2 * X[i] + 2b * w - 2 * w * Y[i]
    )
    DiffOpt.forward_differentiate!(model)
    ∇x[i] = MOI.get(
        model,
        DiffOpt.ForwardVariablePrimal(),
        w,
    )
    MOI.set(
        model,
        DiffOpt.ForwardObjectiveFunction(),
        (2Y[i] - 2b - 2w * X[i]),
    )
    DiffOpt.forward_differentiate!(model)
    ∇y[i] = MOI.get(model, DiffOpt.ForwardVariablePrimal(), w)
end
\end{jllisting}
\end{minipage}

\begin{figure}
  \centering
  \begin{subfigure}[b]{0.48\textwidth}
    \includegraphics[width=\textwidth]{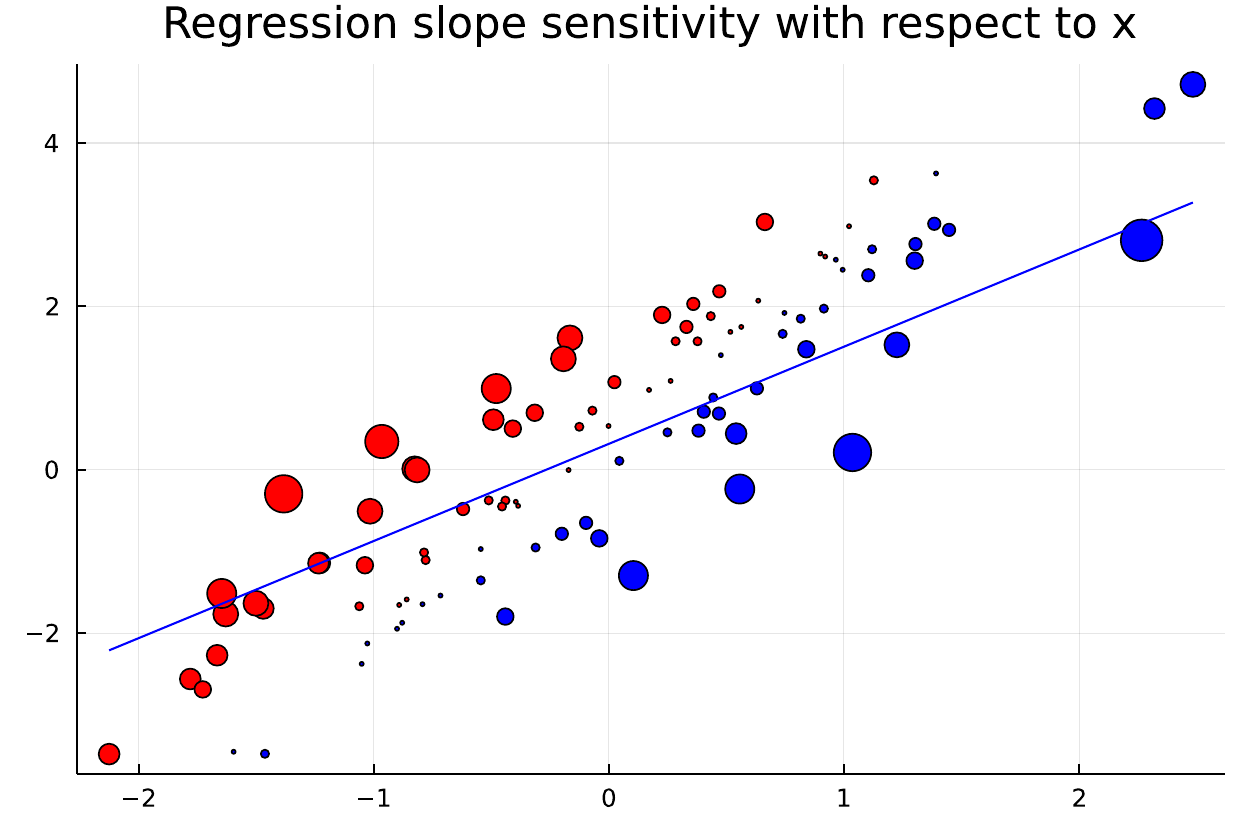}
    \caption{$x$ sensitivity of the regression slope $w$.}
  \end{subfigure}
  \begin{subfigure}[b]{0.48\textwidth}
    \includegraphics[width=\textwidth]{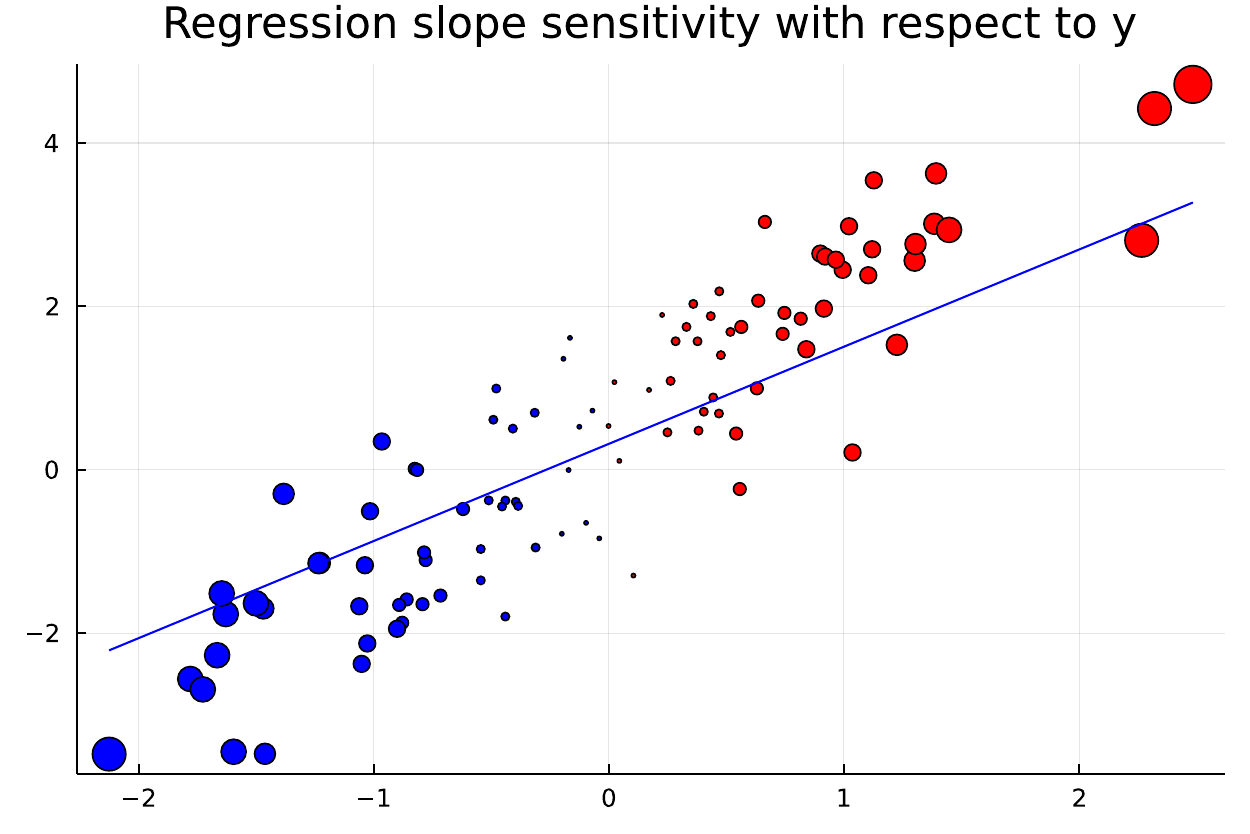}
    \caption{$y$ sensitivity of the regression slope $w$.}
  \end{subfigure}             
  \caption{Sensitivity analysis of the data points in ridge regression. The radius of the markers is proportional to the sensitivity of the slope to x or y perturbations.
  Blue markers indicate a negative sensitivity, red markers a positive one.}
  \label{fig:sens_ridge}
\end{figure}

\noindent
\Cref{fig:sens_ridge} shows how sensitive in amplitude and direction is the slope $w$ to perturbations of each data point.

Similarly to \cref{sec:svm},
the sensitivities can be obtained either using the quadratic or conic DiffOpt models.
The main difference with \cref{sec:svm} is that we now use a conic solver.
The dual solution for the conic constraint is therefore transformed into the dual of the quadratic constraint as detailed in \citet[Example~2.1.2]{legat2020set}.

\subsection{Convex Optimization for Neural Network Layers}\label{sec:convexneuralnet}

Most of the common neural network layers are either simple closed-form operators or a composition
of several operators.
$\partial \mathcal{O}$ opens new possibilities by providing derivatives for
layers defined as solutions to optimization problems.
This example demonstrates such a use case of $\partial \mathcal{O}$, as originally motivated in \citet{amos_differentiable_2019}.
The derivatives computed by DiffOpt
can be used to backpropagate through the optimization layer,
declaring it as a differentiable solution map with \texttt{ChainRules}.
Such an example is already an advanced use case, and the differentiable solution map and its \texttt{ChainRules} derivatives would realistically be defined by an intermediate modeling layer.

\subsubsection{Custom ReLU layer}

This example will follow the tutorial \textit{Custom ReLU layer} of the documentation.
The Rectified Linear Unit or ReLU, a commonly used linear layer in machine learning networks, is defined as $f(x) = \max\{x, 0\}$. It can be interpreted as projecting a point $x \in \mathbb{R}^n$ onto the non-negative orthant:
$y \in \argmin_{y \geq 0} \, \| x - y \|_2^2$ 
where $y$ is the optimization variable and $x$ is the input.

Using the above definition, we model ReLU as a layer in a neural network created in Flux.jl \citep{innes:2018}. The model was trained on MNIST image dataset \citep{lecun2010mnist} with \num{60000} greyscale training bitmaps of size $28 \times 28$. We define the function \texttt{matrix\_relu} with a matrix input because it allows training in batches, the first dimension of the matrix is the layer size, while the second dimension is the size of the batch. We note that, as in the other examples, everything is being executed on CPUs, no GPUs are involved. We use batches in CPUs because we empirically noticed that, for this particular example, it is faster to solve a single large problem than multiple small ones.
\begin{minipage}{\linewidth}
\flushleft
\begin{jllisting}
function matrix_relu(
    y::Matrix;
    model = Model(() -> DiffOpt.diff_optimizer(Ipopt.Optimizer)),
)
    layer_size, batch_size = size(y)
    empty!(model)
    set_silent(model)
    @variable(model, x[1:layer_size, 1:batch_size] >= 0)
    @objective(model, Min, x[:]'x[:] -2y[:]'x[:])
    optimize!(model)
    return value.(x)
end
\end{jllisting}
\end{minipage}

Using our function as a neural network layer requires a corresponding derivative implementation to differentiate the model and propagate the gradients backward. This can be achieved using DiffOpt in conjunction with \texttt{ChainRules} by defining a method for the
reverse-mode primitive function \texttt{rrule}.

\begin{minipage}{\linewidth}
\begin{jllisting}
function ChainRulesCore.rrule(::typeof(matrix_relu), y::Matrix{T}) where {T}
    model = Model(() -> DiffOpt.diff_optimizer(Ipopt.Optimizer))
    pv = matrix_relu(y, model = model)
    function pullback_matrix_relu(dl_dx)
        x = model[:x] # load decision variable `x` into scope
        dl_dy = zeros(T, size(dl_dx))
        dl_dq = zeros(T, size(dl_dx))
        # set sensitivities
        MOI.set.(model, DiffOpt.ReverseVariablePrimal(), x[:], dl_dx[:])
        # compute grad
        DiffOpt.reverse_differentiate!(model)
        # return gradient wrt objective parameters
        obj_exp = MOI.get(model, DiffOpt.ReverseObjectiveFunction())
        # coeff of `x` in q'x = -2y'x
        dl_dq[:] .= JuMP.coefficient.(obj_exp, x[:])
        dq_dy = -2 # dq/dy = -2
        dl_dy[:] .= dl_dq[:] * dq_dy
        return (ChainRulesCore.NoTangent(), dl_dy)
    end
    return pv, pullback_matrix_relu
end
\end{jllisting}
\end{minipage}

\noindent
Note the \texttt{ChainRulesCore.NoTangent} term, which corresponds to the derivative of the output
w.r.t.~the function \texttt{matrix\_relu} itself.
We can now define the neural network architecture including our custom layer and train it on the MNIST data.

\begin{minipage}{\linewidth}
\begin{jllisting}
using MLDatasets
using Flux

# neural network definition
layer_size = 10
m = Flux.Chain(
    Flux.Dense(784, layer_size), # 784 being image linear dimension (28 x 28)
    matrix_relu,
    Flux.Dense(layer_size, 10), # 10 being the number of outcomes (0 to 9)
    Flux.softmax,
)

# dataset preprocessing
N = 1000 # batch size
imgs = MLDatasets.MNIST.traintensor(1:N)
labels = MLDatasets.MNIST.trainlabels(1:N)
train_X = float.(reshape(imgs, size(imgs, 1) * size(imgs, 2), N))
train_Y = Flux.onehotbatch(labels, 0:9);
epochs = 50
dataset = repeated((train_X, train_Y), epochs)

# optimization of the neural network
custom_loss(x, y) = crossentropy(m(x), y) 
opt = Flux.ADAM()
Flux.train!(custom_loss, params(m), dataset, opt) 
\end{jllisting}
\end{minipage}

\noindent
Implementing a custom layer for a known closed-form function is not directly useful, and solving a
quadratic problem is costly in contrast with a simple ReLU operation.
However, it opens the door to more flexible variations of the layer.

\subsubsection{Polyhedral projection layer}

We generalize the custom layer defined as an optimization problem from the ReLU example.
This use case is available in the tutorials as \textit{Polyhedral QP layer}.
Given $m$ vector-scalar pairs $(w_i, b_i)\,\, \forall i \in 1..m$, we define the layer taking $y$ as input
and projecting it on the polytope defined by the $m$ hyperplanes:
\begin{align*}
\min_{x}\,\, & \|x - y\|^2_2 \\
\text{s.t.}\,& w_i^\top x \geq b_i \,\,\, \forall i \in 1..m.
\end{align*}

\noindent
Instead of a function, we will represent the layer with a functor (or callable object).

\begin{minipage}{\linewidth}
\begin{jllisting}
struct Polytope{N}
    w::NTuple{N, Vector{Float64}}
    b::Vector{Float64}
end

Polytope(w::NTuple{N}) where {N} = Polytope{N}(w, randn(N))
\end{jllisting}
\end{minipage}

We define a ``call'' operation on the polytope, making it a so-called functor.
Calling the polytope with a matrix \texttt{y} operates an Euclidean projection of each of the matrix columns onto the polytope.

\begin{minipage}{\linewidth}
\begin{jllisting}
function (polytope::Polytope{N})(
    y::AbstractMatrix;
    model = direct_model(DiffOpt.diff_optimizer(Ipopt.Optimizer)),
) where {N}
    layer_size, batch_size = size(y)
    empty!(model)
    @variable(model, x[1:layer_size, 1:batch_size])
    @constraint(model,
        greater_than_cons[idx in 1:N, sample in 1:batch_size],
        dot(polytope.w[idx], x[:, sample]) ≥ polytope.b[idx]
    )
    @objective(model, Min, dot(x - y, x - y))
    optimize!(model)
    return JuMP.value.(x)
end

Flux.@functor Polytope
\end{jllisting}
\end{minipage}

The \texttt{@functor} macro from Flux implements auxiliary functions for collecting the parameters of our custom layer
and operating backpropagation. Similarly to the ReLU example, \texttt{ChainRulesCore.rrule} is used to implement the
reverse-mode differentiation of the layer.
\begin{jllisting}
function ChainRulesCore.rrule(
        polytope::Polytope{N},
        y::AbstractMatrix) where {N}
    model = direct_model(DiffOpt.diff_optimizer(Ipopt.Optimizer))
    xv = polytope(y; model = model)
    function pullback(dl_dx)
        layer_size, batch_size = size(dl_dx)
        dl_dx = ChainRulesCore.unthunk(dl_dx)
        #  `dl_dy` is the derivative of `l` wrt `y`
        x = model[:x]
        # grad wrt input parameters
        dl_dy = zeros(size(dl_dx))
        # grad wrt layer parameters
        dl_dw = zero.(polytope.w)
        dl_db = zero(polytope.b)
        # set sensitivities
        MOI.set.(model, DiffOpt.ReverseVariablePrimal(), x, dl_dx)
        # compute grad
        DiffOpt.reverse_differentiate!(model)
        # compute gradient wrt objective function parameter y
        obj_expr = MOI.get(model, DiffOpt.ReverseObjectiveFunction())
        dl_dy .= -2 * JuMP.coefficient.(obj_expr, x)
        greater_than_cons = model[:greater_than_cons]
        for idx in 1:N, sample in 1:batch_size
            cons_expr = MOI.get(model,
                DiffOpt.ReverseConstraintFunction(),
                greater_than_cons[idx, sample])
            dl_db[idx] -= JuMP.constant(cons_expr)/batch_size
            dl_dw[idx] .+= JuMP.coefficient.(cons_expr, x[:,sample])/batch_size
        end
        dself = ChainRulesCore.Tangent{Polytope{N}}(; w = dl_dw, b = dl_db)
        return (dself, dl_dy)
    end
    return xv, pullback
end
\end{jllisting}

Note that the inner pullback returns a \texttt{ChainRulesCore.Tangent} that represents the tangent of a composite type.
This will allow Flux to operate gradient descent on the parameters of the \texttt{Polytope} struct directly.
Similarly to the previous example, we can now build and train the network (we omit other details like dataset preprocessing):

\begin{minipage}{\linewidth}
\begin{jllisting}
layer_size = 20
m = Flux.Chain(
    Flux.Dense(784, layer_size), # 784 being image linear dimension (28 x 28)
    Polytope((randn(layer_size), randn(layer_size), randn(layer_size))),
    Flux.Dense(layer_size, 10), # 10 being the number of outcomes (0 to 9)
    Flux.softmax,
)
Flux.train!(custom_loss, Flux.params(m), dataset, opt)
\end{jllisting}
\end{minipage}

The capacity to embed a convex problem as a neural network layer enables formulations of
SVMs or regression layers with custom constraints, which have direct applications in meta-learning \citep{lee2019meta}.

\subsection{Hyperparameter optimization}

Most machine learning algorithms involve hyperparameters that require tuning to
accelerate the training process and reach good out-of-sample performance,
helping achieve a balance between model variance and bias.
In the past few years, many developments were made in gradient-based methods \citep{maclaurin2015gradient} where researchers typically introduced extra hyperparameters for tuning. Recent interest in automated machine learning (AutoML) has resulted in a resurgence of research in this field \citep{feurer2019hyperparameter}.

When the learning process can be formulated as solving an
optimization problem of which the hyperparameters are parameters,
the problem of out-of-sample optimization can be formulated as a bilevel optimization problem \citep{guyon2019analysis}:
\begin{align*}
\min_{\theta} \,\,\, & f(X_{test}, \hat{w}, \theta)\\
\text{s.t. } & \hat{w} \in \argmin_{w} f(X_{train}, w, \theta),
\end{align*}
with $w$ the learned weights of the prediction model, $X_{train/test}$ the training and testing data, $f$ the loss function and $\theta$ the hyperparameter.
Depending on the expression of $\partial f / \partial \theta$, computing (even local) optima of the bilevel optimization problem can be challenging.
We showcase how DiffOpt can be used to meta-optimize
the weights and the hyperparameters following the DiffOpt tutorial \textit{Auto-tuning Hyperparameters}.

\subsubsection{Optimization problem}

Let
$X=\{(x,y) | x \in \mathbb{R}^d, y \in \mathbb{R} \}$ be the set of $N$ data points of dimension $d+1$. The regularized linear model can be modeled as an optimization problem of the form:
\begin{equation}\label{eq:autotune-ridge}
\min_{w} \quad \frac{1}{2nd} \sum_{i=1}^{n} (y_{i} - w^\top x_{i})^2 + \frac{\alpha}{2d} \| w \|_2^2
\end{equation}
where $w \in \mathbb{R}^d$ are the learned weights and $\alpha$,
the regularization parameter, is the only hyperparameter.
Since the problem is strongly convex, it admits a unique minimum $w^*$. Its implementation in \texttt{JuMP} is given below.

\begin{minipage}{\linewidth}
\begin{jllisting}
import JuMP

function fit_ridge(model, X, y, α)
    JuMP.empty!(model)
    N, D = size(X)
    @variable(model, w[1:D])
    @expression(model, err_term, X * w - y)
    @objective(
        model, Min,
        dot(err_term, err_term) / (2 * N * D) + α * dot(w, w) / (2 * D),
    )
    optimize!(model)
    return w
end
\end{jllisting}
\end{minipage}

\subsubsection{Model differentiation}

We want to find the optimal regularization parameter for the
loss on a test set that was not used to find the optimal weights. 
We will apply gradient descent of the unregularized test loss with respect to $\alpha$, for which we compute its gradient $\frac{\partial l}{\partial \alpha}$ using the chain rule:
\begin{equation*}
\frac{\partial l}{\partial \alpha}(w,\alpha) = \frac{\partial l}{\partial w}(w,\alpha) \, \frac{\partial w}{\partial \alpha}(\alpha),
\end{equation*}
\noindent
where $\nabla_{\alpha} w (w,\alpha)$ is the derivative of
the optimal solution of Problem \eqref{eq:autotune-ridge} w.r.t.~the parameter and can be found using
DiffOpt as follows:

\begin{minipage}{\linewidth}
\begin{jllisting}
function compute_dw_dα(model, w)
    D = length(w)
    dw_dα = zeros(D)
    MOI.set(
        model, 
        DiffOpt.ForwardObjectiveFunction(),
        dot(w, w)  / (2 * D),
    )
    DiffOpt.forward_differentiate!(model)
    for i in 1:D
        dw_dα[i] = MOI.get(
            model,
            DiffOpt.ForwardVariablePrimal(), 
            w[i],
        )
    end
    return dw_dα
end
function d_testloss_dα(model, X_test, y_test, w, ŵ)
    N, D = size(X_test)
    dw_dα = compute_dw_dα(model, w)
    err_term = X_test * ŵ - y_test
    return sum(eachindex(err_term)) do i
        dot(X_test[i,:], dw_dα) * err_term[i]
    end / (N * D)
end
\end{jllisting}
\end{minipage}

\subsubsection{Hyperparameter gradient descent}\label{sec:gdhyper}

The value of $\alpha$ is updated using a fixed-step gradient descent scheme implemented below:
\begin{minipage}{\linewidth}
\begin{jllisting}
function descent(α0, max_iters=100; fixed_step = 0.01, grad_tol=1e-3)
    αs = Float64[]
    ∂αs = Float64[]
    test_loss = Float64[]
    α = α0
    N, D = size(X_test)
    model = Model(() -> DiffOpt.diff_optimizer(Ipopt.Optimizer))
    for iter in 1:max_iters
        w = fit_ridge(model, X_train, y_train, α)
        ŵ = value.(w)
        err_term = X_test * ŵ - y_test
        ∂α = d_testloss_dα(model, X_test, y_test, w, ŵ)
        push!(αs, α)
        push!(∂αs, ∂α)
        push!(test_loss, norm(err_term)^2 / (2 * N * D))
        α -= fixed_step * ∂α
        if abs(∂α) ≤ grad_tol
            break
        end
    end
    return αs, ∂αs, test_loss
end
\end{jllisting}
\end{minipage}

\subsubsection{Numerical results}

\begin{figure}
\centering
\begin{minipage}{0.48\textwidth}
\includegraphics[width=\textwidth]{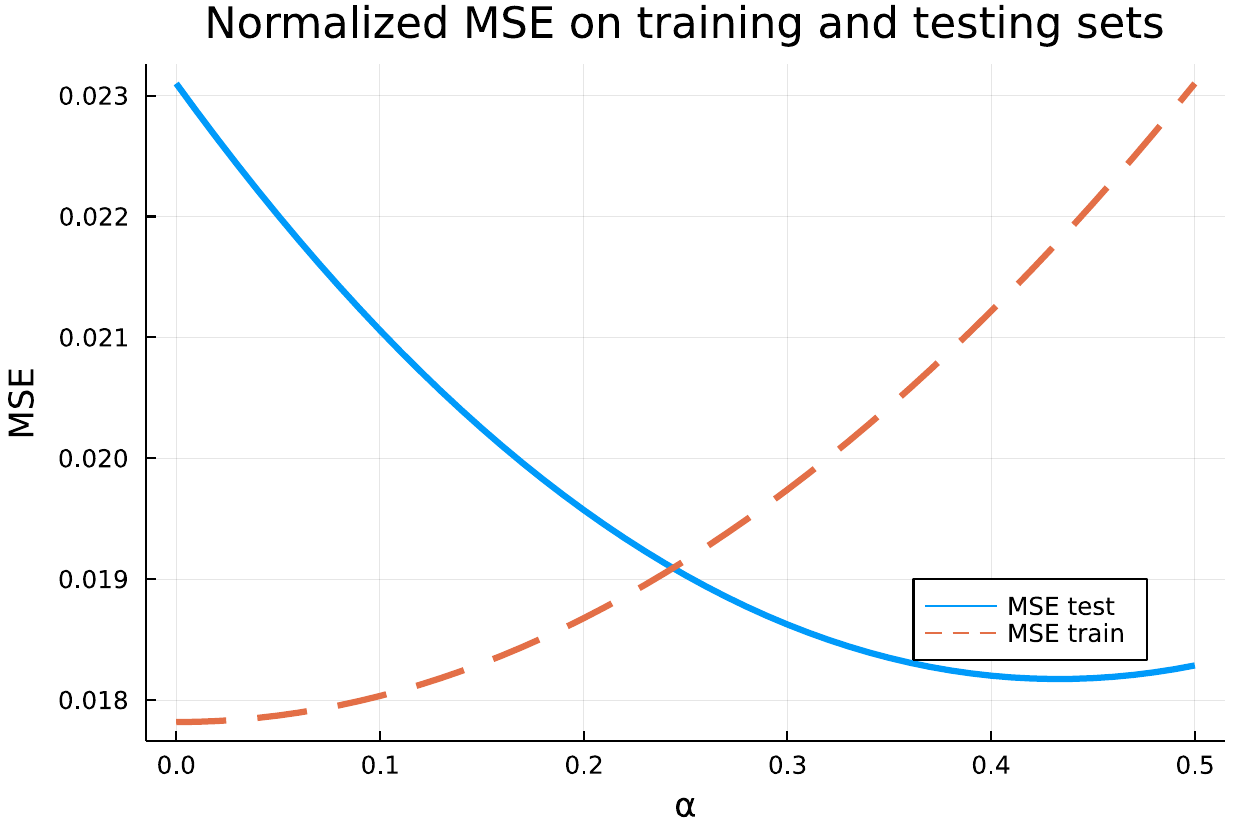}
\caption{\small Mean squared error on the training and test set against $\alpha$}
\label{fig:msetraintest}
\end{minipage}
\begin{minipage}{0.48\textwidth}
\includegraphics[width=\textwidth]{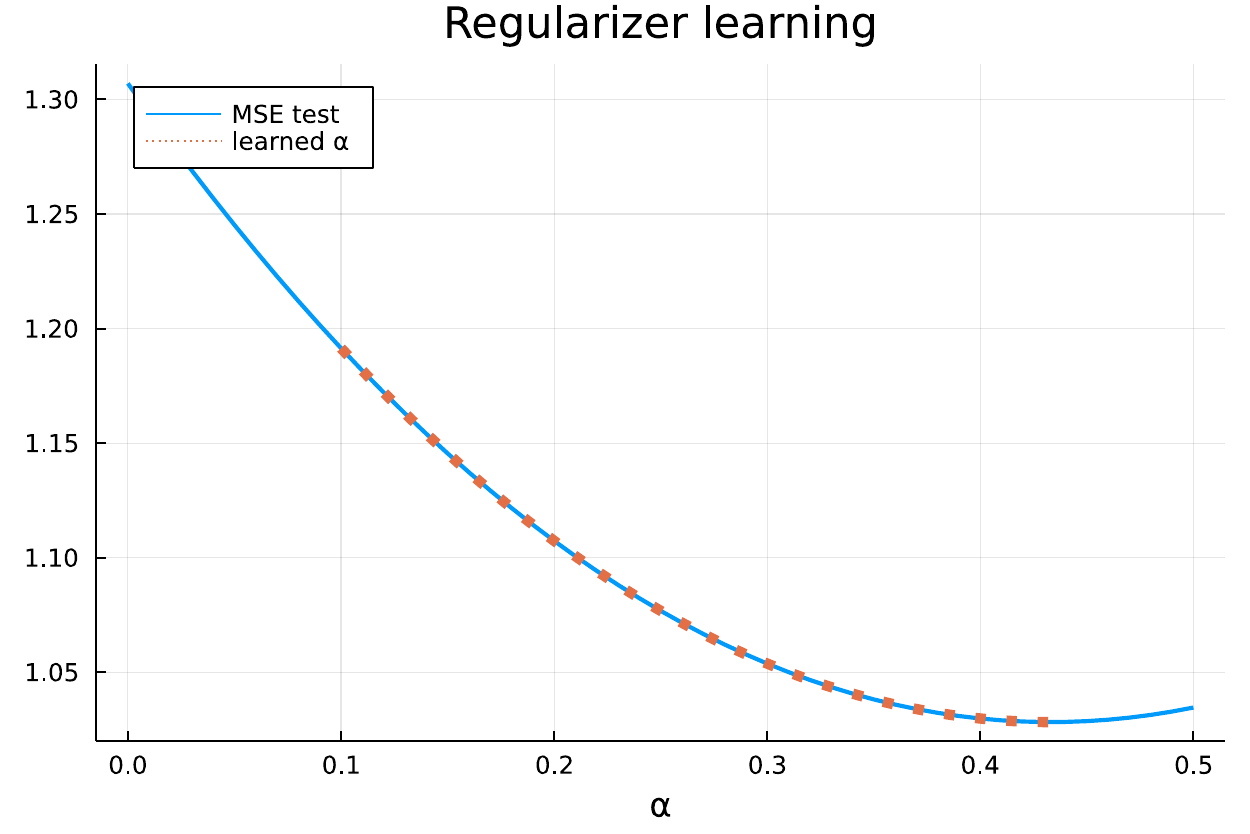}
\caption{\small Gradient descent on the test set for $\alpha$ using DiffOpt}
\label{fig:gradientdescent}
\end{minipage}
\end{figure}

The mean squared error of the regression on the training and test sets
is displayed in \cref{fig:msetraintest}, normalized for display.
As no DiffOpt model is specified, the QP differentiation rules described in \cref{sec:qp} are used. For values of $\alpha$ below $\approx 0.23$, the model is under-regularized, i.e., overfitted to the training data. Increasing $\alpha$ improves the
test error. After that point, increasing $\alpha$ excessively shrinks the
regression coefficients, increasing the error.

\cref{fig:gradientdescent} displays the trajectory of the gradient descent
procedure described in \cref{sec:gdhyper}, starting from $\alpha_0 = 0.1$
and performing about 120 iterations.

Cross-validation is the typical procedure applied to tune the hyperparameters of learning models.
Leveraging $\partial \mathcal{O}$ lets us replace black-box optimization procedures that are typical for hyperparameters
with any first-order method that leverages the gradient obtained from DiffOpt.

\subsection{Nearest correlation matrix}

In finance, the correlation matrix is often only known approximately due to missing data and this approximation may fail to be positive semidefinite.
As studied in \cite{higham2002computing,anjos2003semidefinite,malick2004dual},
the projection of this matrix onto a valid correlation matrix is a weighted least-squares minimization problem on the \emph{elliptope}~\cite{deza1997geometry} i.e.~the set of valid correlation matrices:
\begin{align*}
\min_{X \in \mathbb{S}_+^n} \,\, \| H \odot (X - A) \|_F^2 \,\,\text{ s.t. } X_{ii} = 1 \,\forall i \in 1..n,
\end{align*}
where $\odot$ is the element-wise multiplication, $\| \cdot \|_F$ is the Frobenius norm of a matrix and $\mathbb{S}_+^n$ is the set of positive semidefinite matrices with side dimension $n$.

Other examples of semidefinite least-squares problems include preconditioning of linear systems and error analysis of Jacobi methods for the symmetric eigenvalue problem~\citep{davies2000numerically}.
Consider the distance between a given matrix and its projection onto a closed convex cone $\mathcal{K}$,
\citet[Theorem 2.2]{malick2004dual} shows that the gradient of this distance with respect to the matrix to project $A$
is given by its projection onto the \emph{polar cone} of $\mathcal{K}$.
However, the derivative with respect to the least-squares weights $H$ has not been explored to the best of our knowledge.

Because of the semidefinite constraint, the model cannot be formulated as a quadratic program.
However, thanks to the ability to differentiate through
the second-order cone reformulation of the least-squares objective developed in \cref{sec:quad_to_soc},
we can differentiate it using the conic differentiation rules detailed in \cref{sec:conic}.
We show below how DiffOpt obtains the differentiation with respect to the least-squares weights $H$ of the objective function as discussed in \cite[(1.3)]{higham2002computing}.

\begin{minipage}{\linewidth}
\begin{jllisting}
function projection(A, H, dH)
    n = LinearAlgebra.checksquare(A)
    model = Model(() -> DiffOpt.diff_optimizer(solver))
    @variable(model, X[1:n, 1:n] in PSDCone())
    @constraint(model, [i in 1:n], X[i, i] == 1)
    @objective(model, Min, sum((H .* (X - A)) .^ 2))
    MOI.set(
        model,
        DiffOpt.ForwardObjectiveFunction(),
        sum((dH .* (X - A)) .^ 2),
    )
    optimize!(model)
    DiffOpt.forward_differentiate!(model)
    dX = MOI.get.(model, DiffOpt.ForwardVariablePrimal(), X)
    return value.(X), dX
end
\end{jllisting}
\end{minipage}

\section{Conclusion}

DiffOpt allows users to differentiate through optimization
problems implemented using \linebreak
\texttt{MathOptInterface}, formulated with high-level 
modelling systems like \texttt{JuMP} or\linebreak \texttt{Convex.jl}.
The package brings a flexible differentiable optimization framework
beyond disciplined convex optimization, separating the underlying 
implementation from the MOI idiomatic interface.
This separation of concerns opens the possibility for the integration
of novel differentiable optimization techniques for classes of problems
not already covered. As a first example, we highlight integer programming for which there is still not a mature method to be implemented. A second case is the one of nonlinear programming for which there is a sound theoretical background, but we lack support on MOI which does not yet represents nonlinear constraints as first-class citizens. Once this challenge is overcome, it would be feasible to create a \texttt{NonlinearDiffProblem} \textit{DiffOpt model} analogous to what we discussed in Section \ref{sec:rules}.

Future work will also consider performance improvements.
In particular, using DiffOpt to tune hyperparameters or construct
a convex optimization layer requires fast computation of derivatives
and reoptimization of a perturbed primal problem. The two steps could be
integrated to cache more information and reduce the computational burden per iteration. Finally, solvers like OSQP and sIpopt, which have special versions supporting differentiable optimization natively, will be able to implement the interface defined by DiffOpt and expose the functionality once the low-level Julia wrappers are ready for it.

\section*{Acknowledgments}

The work of A.~Sharma on \texttt{DiffOpt.jl} was funded by the
Google Summer of Code program through NumFocus.
M.~Besançon was partially supported through the Research Campus Modal funded by the German Federal Ministry of Education and Research (fund numbers 05M14ZAM,05M20ZBM).
J.~Dias Garcia was supported in part by the Coordenação de Aperfeiçoamento de Pessoal de Nível Superior - Brasil (CAPES) - Finance Code 001.
B.~Legat was supported by a BAEF Postdoctoral Fellowship and the NSF grant OAC-1835443.
The authors would like to thank all contributors and users of the package for their feedback and improvements, and in particular, Frames Catherine White, Will Tebbutt, and Raphael Saavedra for support, feedback on the API and documentation, and the integration with ChainRules.
We would also like to thank Andreas Varga for his help in solving the matrix equations~\eqref{eq:BXA}.
We thank Guillaume Dalle for the discussion on an early version of this manuscript, and the anonymous reviewers for valuable feedback during the revision process.

\bibliographystyle{apalike}
\bibliography{refs}

\end{document}